\documentclass{article} 
\usepackage{collas2024_conference,times}
\usepackage{easyReview}

\usepackage{amsmath,amssymb}
\usepackage{booktabs}
\usepackage{wrapfig}
\usepackage{algorithm,algpseudocode}

\usepackage{amsmath,amsfonts,bm}

\def\eqref#1{equation~\ref{#1}}

\def\1{\bm{1}}

\DeclareMathAlphabet{\mathsfit}{\encodingdefault}{\sfdefault}{m}{sl}
\SetMathAlphabet{\mathsfit}{bold}{\encodingdefault}{\sfdefault}{bx}{n}

\usepackage{hyperref}
\hypersetup{
    colorlinks=true,
    linkcolor=red,
    filecolor=magenta,
    urlcolor=blue,
    citecolor=purple,
    pdftitle={Overleaf Example},
    pdfpagemode=FullScreen,
    }

\makeatletter
\newcommand{\printfnsymbol}[1]{%
  \textsuperscript{\@fnsymbol{#1}}%
}

\title{Can We Break Free from Strong Data Augmentations in Self-Supervised Learning?}

\author{Shruthi Gowda\textsuperscript{\rm 1}, Elahe Arani\thanks{Contributed equally.}\textsuperscript{\rm~~,1,2}, Bahram Zonooz\printfnsymbol{1}\textsuperscript{\rm,1} \\
\textsuperscript{\rm 1}Department of Mathematics and Computer Science, Eindhoven University of Technology, The Netherlands\\
\textsuperscript{\rm 2} Wayve Technologies Ltd., London, United Kingdom\\
\footnotesize\texttt{\{s.gowda, e.arani, b.zonooz\}@tue.nl}
}

\preprintcopy 

\begin{document}

\maketitle

\begin{abstract}

Self-supervised learning (SSL) has emerged as a promising solution for addressing the challenge of limited labeled data in deep neural networks (DNNs), offering scalability potential. However, the impact of design dependencies within the SSL framework remains insufficiently investigated. In this study, we comprehensively explore SSL behavior across a spectrum of augmentations, revealing their crucial role in shaping SSL model performance and learning mechanisms. Leveraging these insights, we propose a novel learning approach that integrates prior knowledge, with the aim of curtailing the need for extensive data augmentations and thereby amplifying the efficacy of learned representations. Notably, our findings underscore that SSL models imbued with prior knowledge exhibit reduced texture bias, diminished reliance on shortcuts and augmentations, and improved robustness against both natural and adversarial corruptions. These findings not only illuminate a new direction in SSL research, but also pave the way for enhancing DNN performance while concurrently alleviating the imperative for intensive data augmentation, thereby enhancing scalability and real-world problem-solving capabilities.
\footnote{The code is available at: \url{https://github.com/NeurAI-Lab/SSL-Prior}.}



\end{abstract}

\section{Introduction}
\setlength\intextsep{2pt}
\begin{wrapfigure}{R}{0.525\textwidth}
    \centering
    \includegraphics[width=0.515\textwidth]{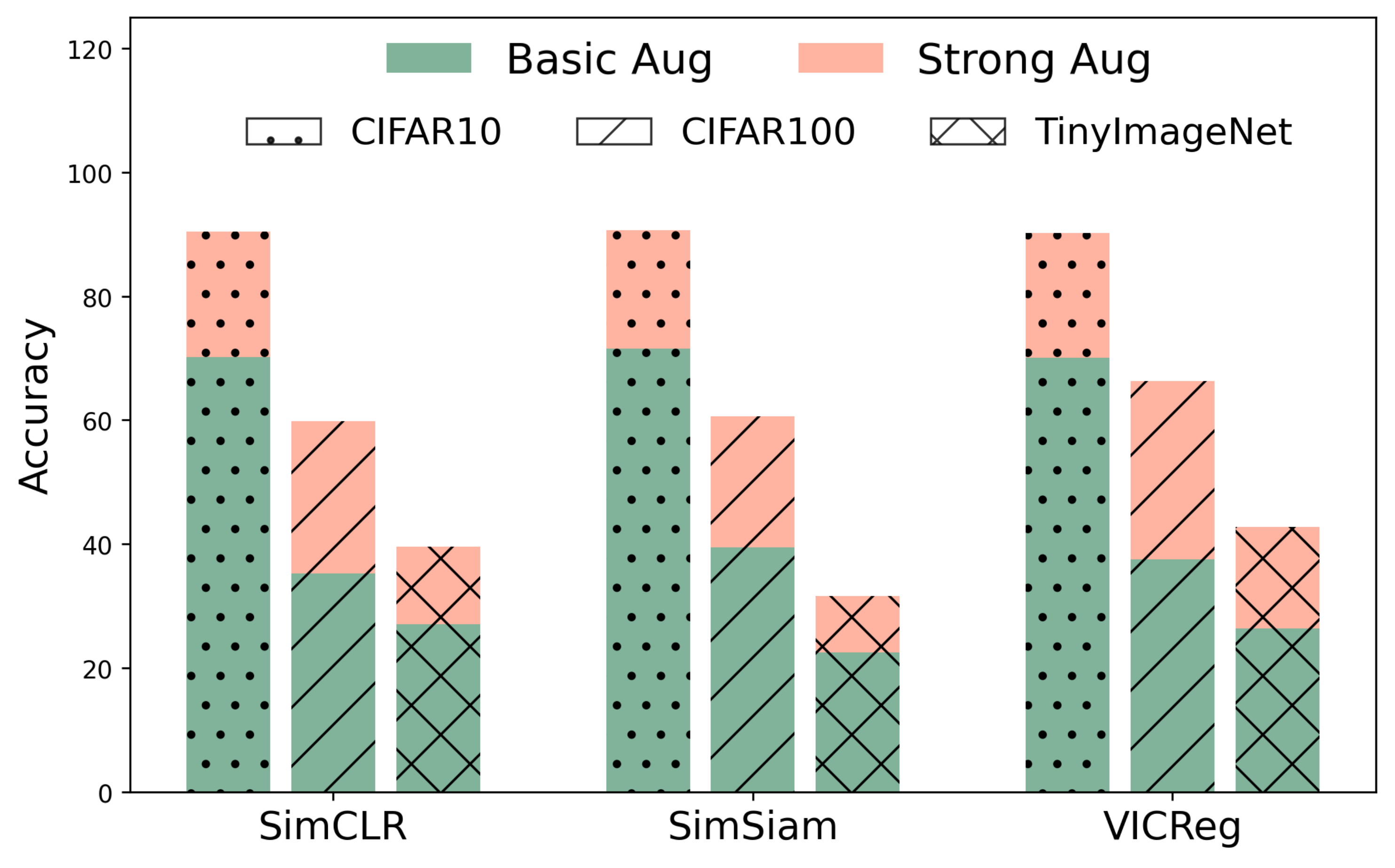}
    \caption{\label{fig:first}The impact of augmentations on SSL methods is critical: as removing strong augmentations from SSL training can result in a significant drop in their performance.}
\end{wrapfigure}
Deep neural networks (DNNs) have proven to be highly effective in encoding patterns in data distribution to produce powerful and rich representations that have improved generalization performance across various perception tasks, such as classification, detection, and segmentation. However, one of the major limitations is that DNNs are data-hungry and annotating millions of available data is expensive. Self-supervised learning (SSL) has been proposed as a promising solution to this issue, to enable the learning of useful representations without manual annotations. Self-supervised learning paradigm 
needs to ensure that the resulting features are generic to be applicable to a wide range of real-world applications.


Various SSL methods, including pretext-based \citep{gidarisunsupervised, noroozi2016unsupervised}, contrastive-based \citep{chen2020simple, he2020momentum, chen2020improved}, architectural change-based \citep{grill2020bootstrap, chen2021exploring}, and feature decorrelation-based \citep{bardesvicreg, zbontar2021barlow} have been proposed. The main objective of these methods is to learn generic, high-level representations that can be applied to a variety of downstream tasks. However, the unique design strategies and dependencies involved in SSL training, and their potential impacts, remain largely unexplored. One common practice in SSL training is the use of a large set of augmentations. Although augmentations can be a powerful tool in SSL, their potential implications must be carefully considered. \textit{How do different SSL methods perform without strong data augmentations? What are the impacts on shortcut learning and robustness in real-world downstream applications? Is it possible to design alternate solutions that can achieve comparable performance without relying on an intense set of augmentations?} Investigating these questions can provide insight into the capabilities and limitations of SSL methods, thus guiding future improvements.

\begin{figure*}[t]
 \begin{center}
    \includegraphics[width=0.7\linewidth]{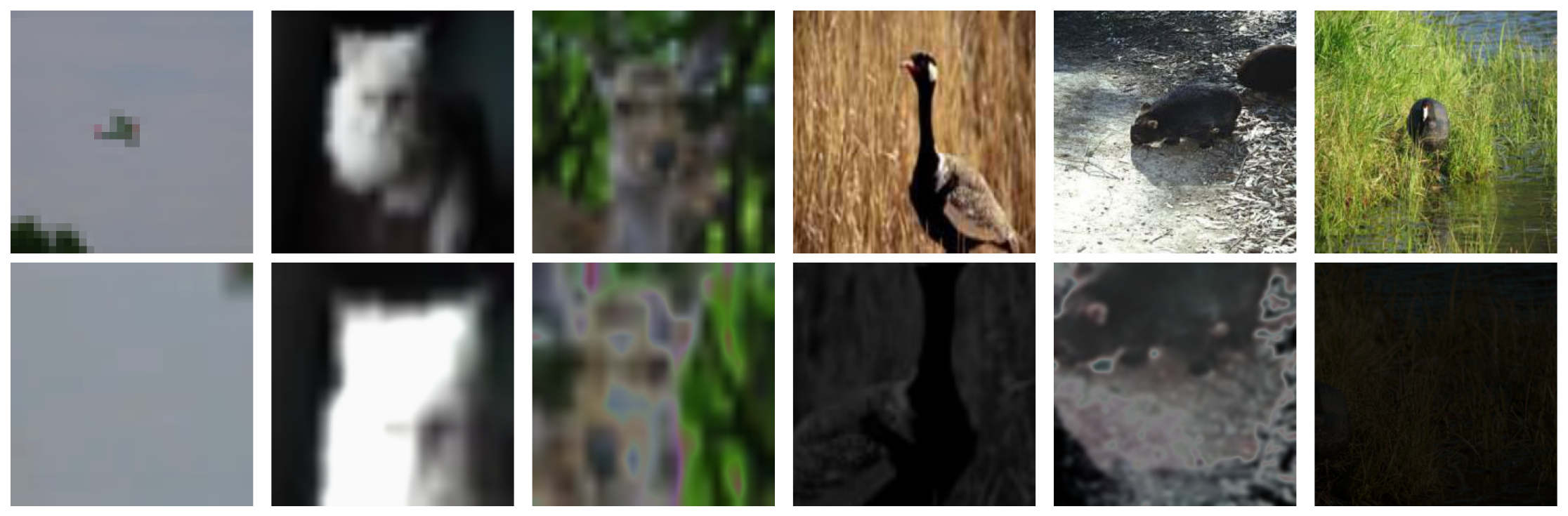}
 \end{center}
  \caption{Examples of aggressive augmentations producing noisy images in CIFAR (first three columns) and ImageNet (last three columns) datasets, respectively. High intensity in augmentations such as color jitter, blurring, and solarization results in semantic shifts.}
  \label{fig:agg_aug}
\end{figure*}

To gain a deeper understanding of the relationship between SSL learning and strong invariances, we evaluated SSL methods on the basic augmentations (from Table \ref{tbl:aug}) and observed a drop in the performance of all methods, showing their dependency on strong augmentations (see Figure \ref{fig:first}). Although strong augmentations enforce multiple invariances in the learned representations, they can also be viewed as a double-edged sword, impacting the downstream application positively or negatively. The intensity of the augmentations can be quite high, altering the object semantics and introducing artificial patterns and correlations into the data \citep{bairsa} (see Figure \ref{fig:agg_aug}) and undesirable biases in the representations \citep{balestriero2022effects}. However, the current direction in SSL research involves proposing new methods that heavily rely on more intense augmentations, and an overdependence on a single factor for performance gain may not necessarily be beneficial in the long term. Therefore, it is crucial to explore alternative directions for improving SSL research.

In contrast to the common pattern observed in SSL methods, which heavily rely on enforced invariances and distortions, the human brain can learn from unlabeled data without such dependencies. This discrepancy suggests that innate cognitive biases may play a significant role in human learning \citep{kuhl2000new}. Motivated by this observation, we propose an alternative approach to SSL that moves away from the current trend of relying on more data augmentations. Specifically, we aim to incorporate prior knowledge, similar to cognitive biases, to enhance the effectiveness of learned representations. By leveraging implicit information that is present in the data, such as shape, we can provide additional supervision and regularization to improve the learned representations. The induction of semantic information into the SSL training paradigm can further mitigate shortcomings, such as texture bias \citep{geirhos2019imagenettrained}, and reduce their reliance on shortcuts and augmentations.

We perform various analyses to evaluate the efficacy of SSL with Priors. Our results show that SSL models with priors are less dependent on augmentations, less susceptible to shortcuts, and exhibit reduced texture bias. SSL models incorporating priors exhibit improved robustness to both natural and adversarial corruptions, demonstrating their ability to generate more generic representations. This trait proves advantageous for unbiased performance in challenging downstream tasks in domains such as autonomous driving and healthcare.
Overall, our study elucidates the behavior of SSL and proposes an alternative approach to elevate the quality of representations and unlock a new era of possibilities in SSL across diverse applications.

\section{Biases in SSL training}


The field of SSL is continuously growing at a fast pace, and there have been recent advancements that effectively use unlabeled data to learn generic representations. However, there are also design choices and dependencies that could potentially have negative impacts. Since the primary goal of SSL is pre-training, it is critical to avoid inducing biases during the learning process, as this could lead to models that are less suitable for downstream applications. Therefore, it is essential to carefully examine the impact of various aspects of SSL training to gain a comprehensive understanding of its capabilities.

\textbf{Pretext tasks.}
Pretext tasks in SSL require the model to predict certain properties or relationships within the data, such as rotation, colorization, or position of an image patch, to capture important features and patterns within the data \citep{gidarisunsupervised, noroozi2016unsupervised}. However, these learned representations heavily depend on the pretext task, and certain features that are useful to solve the task can be easily exploited by adversaries to make it harder for the network \citep{minderer2020automatic}. To address this issue, \citet{minderer2020automatic} propose an additional network that reconstructs perturbed images, which are then sent to the encoder for pretext task training using an adversarial approach. However, these methods do not generalize well to non-pretext based techniques.

\textbf{Strong Augmentations.} 
In SSL, a notable implicit bias is the overreliance on a large number of augmentations to create a diverse set of training examples \citep{balestriero2022effects}. While augmentations are intended to improve the quality and generalization of learned features, their choice and intensity can significantly affect performance. Augmentations can introduce artificial patterns and correlations that bias learned features and hinder their transferability to new or unseen data. Aggressive augmentations can also damage the original information and semantics of the data, hurting the performance of downstream tasks that depend on such information \citep{bairsa}. Therefore, it is essential to balance the usage and intensity of augmentations to maximize their effectiveness and minimize their potential negative impacts.


\textbf{Shortcut Learning.} Shortcut learning is a common issue in deep neural networks (DNNs) that can impact the generalization of the network on new, unseen data. In SSL, shortcut learning can be particularly detrimental, as networks need to learn generic high-level features that can be applied to various downstream tasks. \citet{geirhos2020shortcut} defined shortcut learning as the overreliance on non-causal cues in the input data, which may not be valid for new test data. To address this shortcoming, \citet{robinson2021can} proposed a method to remove the features that are learned easily during the instance discrimination task, forcing the network to learn other features for discrimination. \citet{chen2021intriguing} highlighted the impact of shortcut learning on SSL performance. Despite efforts to address this issue, shortcut learning remains prevalent in DNNs.


\textbf{Texture Bias.} Representation learning often displays texture bias, which is the tendency of deep learning models to rely heavily on local texture information, such as patterns or colors in the data, rather than on higher-level information \citep{geirhos2019imagenettrained, gowda2022inbiased}. This can result in overfitting, poor generalization, and limited transferability of learned features. To mitigate texture bias in networks, \citet{mishra2020learning} propose a new augmentation that suppresses texture information during SSL training. However, solutions beyond augmentations may also be necessary to address texture bias in SSL more comprehensively, and further research is required to mitigate this bias.


\begin{table}[t]
    \begin{center}
    \caption{Basic set of augmentations used in standard classification training and strong set used in SSL training.}
    \label{tbl:aug}
    \begin{small}
    \resizebox{\textwidth}{!}{
    \begin{tabular}{l|llllll}
    \toprule
    \multicolumn{7}{c}{Augmentations} \\ \midrule
\textbf{Basic} & Random Resized Crop & Random Horizontal Flip &  &  &  &  \\ \midrule
\textbf{Strong} & Random Resized Crop & Random Horizontal Flip & Random Grayscale & Color Jitter & Gaussian Blur & Solarization \\ \bottomrule
    \end{tabular}}
    \end{small}
    \end{center}
\end{table}

\begin{figure*}[t]
  \centering
    \includegraphics[width=0.8\linewidth]{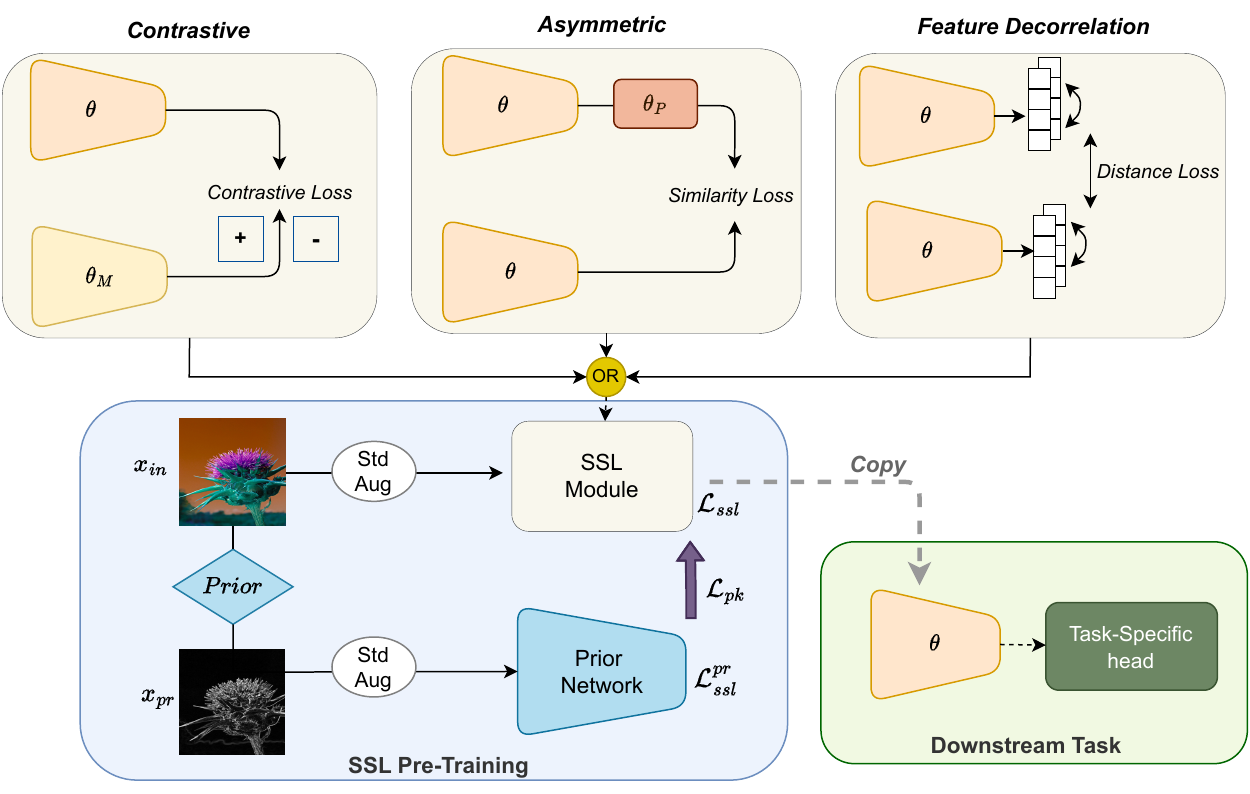}
  \caption{Schematic of SSL method with \textit{Prior knowledge} integration.
  The SSL module can incorporate any SSL method, such as Contrastive, Asymmetric, and Feature Decorrelation-based, and one method is selected from each category for this study. The prior network extracts implicit semantic knowledge and supervises the SSL module network to learn better representations. The resulting network from the SSL module is then used for inference purposes. This approach is expected to improve the quality of learned features and enhance the generalization capability of the resulting network.}
  \label{fig:ssl_method}
\end{figure*}

\section{Background: Self-Supervised Learning}


We categorize SSL methods into three broad categories: contrastive-based, asymmetrical network-based, and feature decorrelation-based. We choose one method from each of these categories and use it for our analyses.

\subsection{Contrastive-based SSL}

Contrastive-based methods aim to learn a representation by maximizing the similarity between different views of the same image and minimizing the similarity between views of different images. SimCLR \citep{chen2020simple} is one such contrastive-based method that has gained significant attention due to its simplicity and effectiveness.

The architecture of SimCLR consists of a base Siamese network, which is useful for finding similarities between inputs. The Siamese network is made up of two identical networks that process two different inputs. The SimCLR encoder $f_\theta$ is appended with a projector $g$ to help project the output of one view to match the other. Given the two views of the input $x_{in}$ as $x$ and $x'$, the encoder outputs for the two augmented views are $h= f(x)$ and $h'= f(x')$, respectively. The projection head takes the encoder outputs and maps them into a shared latent space, where they can be compared. The projection head outputs are $z= g(h)$ and $z'= g(h')$. SimCLR considers one positive pair, and the rest of the data points are considered as negative samples. The contrastive loss objective is provided as,
\begin{equation*}
\mathcal{L}_{ssl}=-\log \frac{\exp \left(\operatorname{sim}\left(\mathbf{z}, \mathbf{z'}\right)\right)}{\sum \mathbb{}_{z \neq z'} \exp \left(\operatorname{sim}\left(\mathbf{z}, \mathbf{z}'\right)\right)}
\end{equation*}

\subsection{Asymmetric Network-based SSL}

Asymmetric network-based SSL approaches avoid mode collapse by introducing architectural asymmetry between the twins of the Siamese networks. SimSiam \citep{chen2021exploring} is one such method that learns representations without the use of negative samples, large batch sizes, or momentum encoders. The SimSiam architecture contains an additional predictor head (an MLP network), and the parameter update is asymmetric, where the network parameters are updated for one augmented version of the input while considering the features of another augmented version as the target.
Given the encoder output: $h= f(x)$; $h'= f(x')$, and the projector output: $z= g(h)$; $z'= g(h')$, SimSiam minimizes the negative cosine similarity $D$ between the predictor output and the projector output as SSL loss:  
\begin{equation*} \label{eq:simsiam1}
    \mathcal{D}\left(y, z\right)=-\frac{y}{\left\|y\right\|_2} \cdot \frac{z}{\left\|z\right\|_2}
\end{equation*}
The SimSiam loss is an asymmetric loss, and the stop-gradient operation is used to calculate both losses. By avoiding the need for negative samples and the use of large batch sizes, SimSiam provides a simpler and more efficient alternative to contrastive-based methods.
\begin{equation*} \label{eq:simsiam}
 \mathcal{L}_{ssl}= \frac{1}{2} {D(y, stopgrad(z'))} + \frac{1}{2} {D(y', stopgrad(z))}
\end{equation*}

\subsection{Feature Decorrelation-based SSL}

Recently, several SSL methods have emerged that eliminate the requirement for negative examples, momentum encoders, or asymmetric mechanisms in the architecture.
VICReg \citep{bardesvicreg} is one such method that has the same basic architecture as its predecessors but has three key terms: variance, invariance, and covariance to regularize training. Variance term encourages the encoder to generate different representations for different views of the same image and aims to promote diversity in the learned representations. Invariance term minimizes the difference between the learned representations of the original input and its augmented versions. Covariance term maximizes the mutual information between the learned representations of two different inputs.
\begin{equation*}
\begin{split}
    \mathcal{L}_{ssl} = \underbrace{\frac{1}{N} \sum_{i} max(0, \gamma - \sqrt{Var(z)})}_{\text{var}} 
    + \underbrace{\frac{1}{N} \sum_{i} ||z_i - z_i'||^2}_{\text{inv}} 
    + \underbrace{\frac{1}{N-1} \sum_{i} (z_i - \mu) (z_i - \mu)^{T}}_{\text{cov}}
\end{split}    
\end{equation*}
where $\mu$ is the mean of the embeddings and $\gamma$ is a target standard deviation.

\section{Proposed Solution}


\textit{How does the behavior of SSL trained models change when intense augmentations are removed?}
By investigating these questions, we gain insights into the robustness and limitations of SSL methods. To evaluate the impact of intense augmentations, we consider a simplified set of basic augmentations derived from a list of common SSL augmentations shown in Table \ref{tbl:aug}. Figure \ref{fig:first} displays the baseline accuracies of SSL models trained with strong and basic augmentations. The results show a significant drop in SSL baseline accuracies when strong and high-intensity augmentations are removed. 

\textit{Can we train SSL methods with less dependency on strong augmentations?}
The current trend in SSL involves using more and intense augmentations to achieve higher performance. However, we aim to explore an alternative approach that can work with basic augmentations while still improving downstream performance. Our approach draws inspiration from the brain, where unlabeled data is also learned without any invariances or operations. We propose incorporating a cognitive bias to leverage this knowledge and enable networks to move beyond simplistic solutions in the data and enhance generalization capabilities.


\textbf{Inductive bias} refers to the inherent preferences or assumptions that guide the learning process in neural networks. It is largely innate and can arise from the structure and function of the brain, as well as from prior experience and learning \citep{pearl2018book}. In neural networks, inductive bias can take many forms, such as using convolutional layers to enforce translational invariance or recurrent networks to incorporate temporal dependencies in sequential data. In this study, we propose inducing another form of inductive bias through semantic information supervision. 

To achieve this, we integrate prior knowledge into the self-supervised training framework to enhance the network's ability to learn better representations. A psychophysical experiment \citep{geirhos2019imagenettrained} demonstrated that the learning patterns of networks and humans differ, with networks relying heavily on textures while humans rely more on shape features. Therefore, we use the "shape" as our template and extract this information, implicit in the data, to provide supervision to the SSL learning technique and encode high-level abstractions. The way this knowledge is induced is of high importance as just adding shape as another augmentation leads to sub-optimality (\ref{sec:abl_app}). Rather than adding a new augmentation, our objective is to supervise the network to learn to look at global structures and improve representations.

\begin{table}[t]
\caption{Comparison of linear classification accuracy on IID and OOD datasets using three SSL methods, all trained on Basic augmentations. Prior knowledge helps the SSL modules learn more generic features and shows consistent improvement against all the baselines.}
\label{tbl:acc}
\begin{center}
\begin{tabular}{l|cccc|cc}
\toprule
& \multicolumn{4}{c|}{IID} & \multicolumn{2}{c}{OOD} \\ \midrule
& CIFAR10 & CIFAR100 & STL10  & TinyImageNet & STL10 & ImageNet-B \\ \midrule
SimCLR  & 70.02 & 35.21 & 67.03 & 27.01 & 35.48 & 10.15 \\
SimCLR-Prior & \textbf{87.45} & \textbf{58.33} & \textbf{82.24} & \textbf{40.12} & \textbf{42.12} & \textbf{12.50} \\ \midrule
SimSiam & 70.30 & 36.78 &  59.38 & 22.47 & 33.15 & 7.29 \\
SimSiam-Prior & \textbf{87.30}  & \textbf{66.06} & \textbf{ 77.56} & \textbf{33.45} & \textbf{39.06} & \textbf{8.10} \\ \midrule
VICReg & 71.56 & 37.50 & 64.21 & 26.37 & 36.32 & 10.41 \\
VICReg-Prior & \textbf{86.81} & \textbf{ 60.22} & \textbf{80.12} & \textbf{35.00} & \textbf{40.18} & \textbf{10.85} \\ 
\bottomrule
\end{tabular}
\end{center}
\end{table}


\subsection{Prior Knowledge Induction}



In our proposed framework, SSL-Prior, we incorporate an SSL module and an additional prior network to induce prior knowledge and improve the representations learned by the SSL module. Specifically, we train the prior network to learn shape information as an additional view by treating it as a separate input. The SSL module can be any existing SSL method, and in this study, we choose one method from each of the three categories: SimCLR, SimSiam, and VICReg. The SSL module typically consists of an encoder network $f$, a projector network $g$, and a predictor $p$, with the specific architecture depending on the chosen SSL method.

To generate the inputs for the SSL module, we perform augmentations $T$ and $T'$ on the original sample $x_{in}$, resulting in two views $x$ and $x'$. The prior filter, which is a Sobel shape filter, is then applied to the input $x_{pr} = Prior(x_{in})$ to extract the shape information before being fed into the prior network. By training these two modules together with aligned objectives, we can reduce overfitting and encourage the network to explore the semantic space and learn more generic representations. For more details on the Sobel shape filter and other implementation specifics, refer to Appendix.



We leverage the rich shape representations produced by the prior network to supervise the SSL module. Specifically, we train the prior network using the same objective as the SSL module, which can be any of the contrastive loss, cosine similarity loss, or variance-invariance-covariance loss. In addition to the SSL-based objective, we transfer prior knowledge through a knowledge transfer objective, where we introduce a consistency loss to distill shape knowledge by computing the Kullback-Leibler divergence $\mathcal{D}_{KL}$ between the network outputs. By aligning the distributions, the network is guided towards less susceptible decisions and less reliant on trivial solutions. The shape view serves as a regularization term, enforcing the network to learn to incorporate global semantic information, while reducing the need for excessive augmentations to achieve invariances.
\begin{equation}
    \mathcal{L}_{pk} = \mathcal{D}_{KL}(\text{softmax}(z) || \text{softmax}(z'))
\end{equation}


The overall loss is the SSL loss along with the prior knowledge loss. The SSL loss learns invariant representations between two views of standard images, while the SSL-Prior loss regularizes the features by maintaining consistency across an additional global semantic view,
\begin{equation}
    \mathcal{L} =  \mathcal{L}_{ssl} + \lambda \mathcal{L}_{pk}
 \label{eq:total-loss}
\end{equation}
where $\lambda$ is the loss balancing factor. Note that $\mathcal{L}^{pr}_{ssl}=\mathcal{L}_{ssl}$.

\begin{figure*}[t] 
    \centering
    \includegraphics[width=0.8\textwidth]{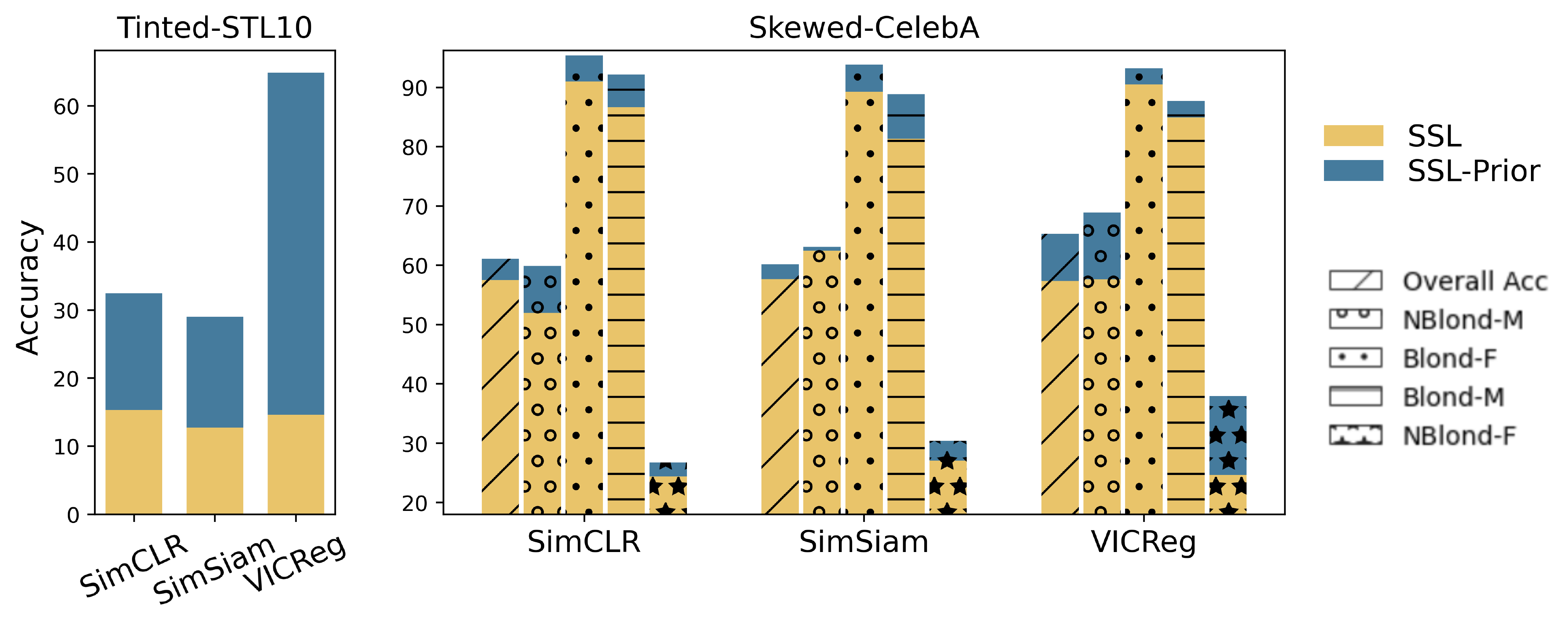}
    \caption{Shortcut learning: Evaluation using Tinted-STL10 and Skewed-CelebA datasets on all three SSL methods. The results indicate that SSL trained with priors are less vulnerable to learning unintended cues and spurious correlations in the data.}
    \label{fig:shortcut}%
\end{figure*}

\section{Experimental Setup}

We categorize SSL into three broad categories and choose one method from each for our study. SimCLR from contrastive-based, SimSiam from asymmetric-based, and VICReg from the feature decorrelation-based approaches.

ResNet-18 architecture \citep{he2016deep} is used in all experiments. A two-layer projection MLP layer is appended to the encoder with a hidden dimension of 2048. For asymmetric methods that require a predictor, there is an additional two-layer MLP predictor. The prior network also consists of a ResNet-18 architecture, and a shape filter (Sobel filter) is incorporated to preprocess and extract shape information from the input image. SSL training is performed using Adam optimizer and Cosine Annealing scheduler. List of hyperparameters for each method is provided in Appendix. 


The self-supervised pre-training is performed on the dataset without using labels. For evaluation, a fine-tuning step is performed where a linear classifier head is trained on top of a frozen encoder. This encoder and linear classifier is used to evaluate on the validation test set to report the accuracy. The encoder $f$ is used for all inference purposes.

\begin{figure*}[t] 
    \centering
    \includegraphics[width=0.85\textwidth]{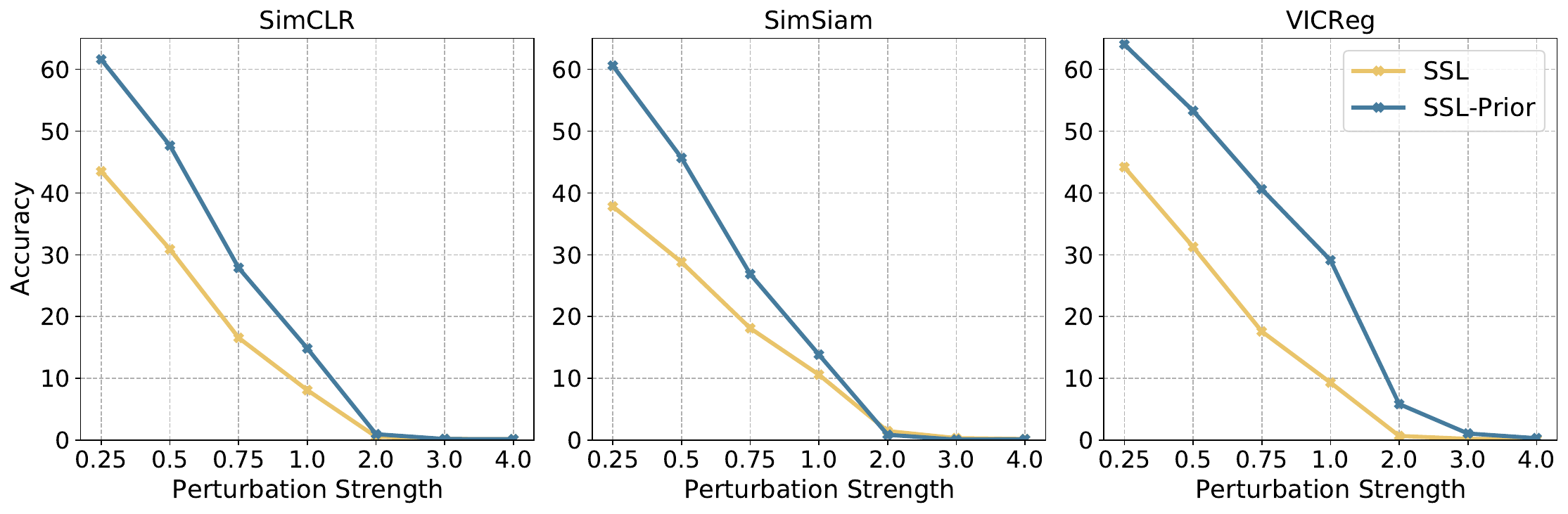} 
    \caption{Robustness analysis: PGD attack on models fine-tined on CIFAR10 dataset. The SSL methods with Prior are more robust compared to the baseline across varying attack strengths.}
    \label{fig:adv}%
\end{figure*}

\section{Results}

\subsection{IID Performance}


Table \ref{tbl:acc} shows the IID performance of the three SSL baselines and their SSL-Prior counterparts for four different datasets. As previously observed in Figure \ref{fig:first}, the SSL baselines exhibit significantly reduced performance when trained with a basic set of augmentations. SSL-Prior, in contrast, demonstrates strong performance across all of the datasets. Furthermore, the SSL-Prior results are comparable to the SSL baselines trained with a strong set of augmentations (as shown in the Appendix).
By incorporating shape prior knowledge, our SSL-Prior approach is able to generate more high-level abstractions and reduce the reliance on intense augmentations. Thus, this alternate solution has the potential to steer the training paradigm toward a new direction of SSL training.

\subsection{OOD Performance}

One of the objectives of SSL is to learn generic representations that can be transferred to other data distributions, without depending on specific tasks or labels. Hence, we test the generalization capability to distribution shift by evaluating the models on out-of-distribution data. 
We perform out-of-distribution (OOD) analysis on two datasets. We utilize our model trained on CIFAR10 to test on the STL10 dataset and the model trained on TinyImageNet to test on Imagenet-B \citep{hendrycks2021many} dataset.
Imagenet-B contains the blurry version of the ImageNet dataset.
As presented in Table \ref{tbl:acc}, SSL-Prior models exhibit significantly improved performance even on challenging OOD scenarios. The SSL-Prior approach can help mitigate the issue of distribution shift and improve the overall generalization ability of the learned representations.


\subsection{Shortcut Learning}

The behavior of DNNs to rely on shortcuts is undesirable as these spurious associations might not translate to the unseen data in the downstream tasks \citep{xiao2021noise}. To evaluate this behavior, we present two classification use cases. For analysis, we utilized models pre-trained on the TinyImageNet dataset and fine-tuned the classifier head on the respective datasets for evaluation.

\textbf{Tinted-STL10} \citep{jain2022combining} is a dataset that is created by adding a class-specific tint to the original STL10 dataset, and the goal is to test whether the models rely on the easy color-class correlation in the data or look beyond this spurious cue to be more generic. The classifier model fine-tuned on this spurious dataset is tested on the original STL10 dataset. 

\textbf{Skewed-CelebA} \citep{jain2022combining} is a dataset that is created by selecting only `blond-females' (Blond-F) and `non-blond-males' (Non-Blond-M) samples from the gender classification dataset CelebA \citep{liu2015deep}. The test set comprises all possible combinations of hair color and gender.

Figure \ref{fig:shortcut} clearly demonstrates that SSL techniques rely heavily on spurious correlations and induced unintended features, such as tint or hair color, which makes them less effective on test settings. This behavior is consistent even when SSL models are trained with the complete set of augmentations (Section \ref{sec:short_app}). In contrast, our SSL-Prior approach exhibits improved robustness to shortcut learning in both use cases. By incorporating shape prior knowledge, the networks are able to learn more semantic information, which reduces the reliance on unintended cues like tint in Tinted-STL10. In Skewed-CelebA, the prior help the models to focus on salient facial features for classification, instead of relying on hair color. This is particularly evident in the exceptional performance of SSL-Prior on the Non-Blond-F category, which was not present in the training data. Thus, the supervision from shape helps the network to focus more on relevant features, rather than trivial solutions.



\begin{figure}[t] 
    \centering
    \includegraphics[width=\textwidth]{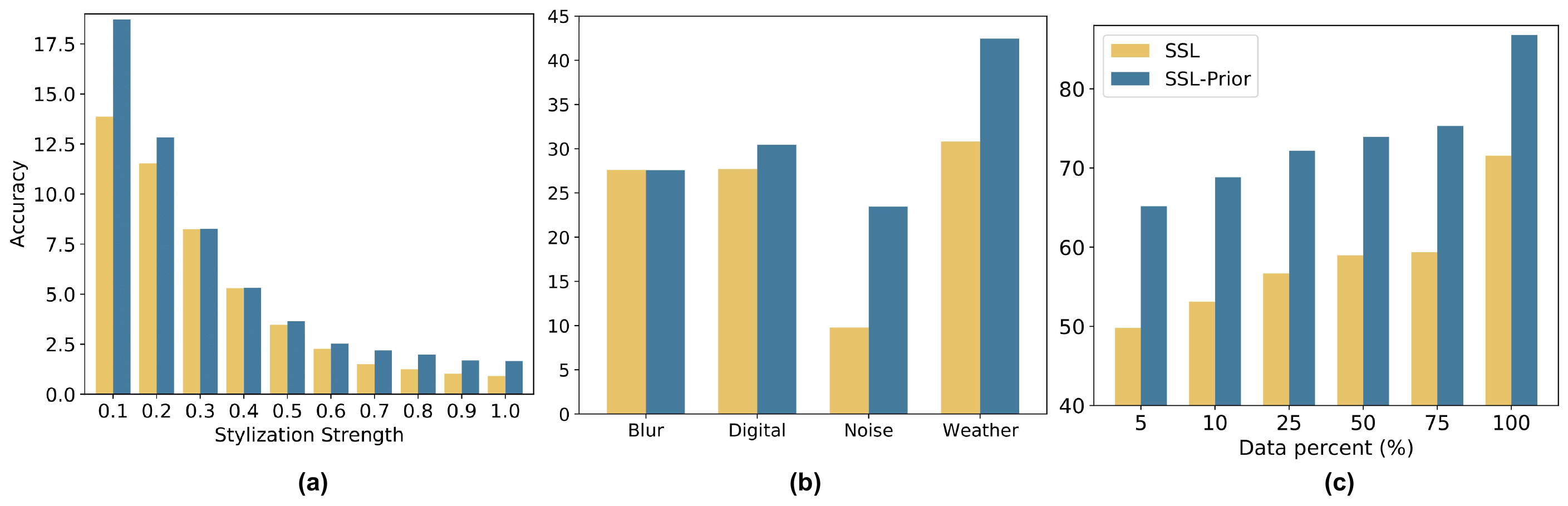}
    \caption{(a) \textbf{Texture Bias:} The SSL method (VICReg) and evaluated on stylized TinyImageNet. The addition of prior knowledge leads to reduced texture bias. (b) \textbf{Robustness:} The SSL method (VICReg) fine-tuned on CIFAR10, evaluated on natural corruptions. Shape awareness helps especially in noisy and changing weather conditions. (c) \textbf{Data Impact:} Performance of SSL model (VICReg) fine-tuned on different percentages of CIFAR10 data.}
    \label{fig:combined}%
\end{figure}

\subsection{Robustness against Adversarial Attacks}

The SSL-trained networks are applicable in various real-time downstream applications in challenging scenarios, where robustness plays a critical role. However, DNNs are vulnerable to adversarial attacks, where even small perturbations to input data can lead to significant misclassification errors. To evaluate the robustness of SSL-trained models, we fine-tuned the pre-trained models on TinyImageNet for classification on the CIFAR10 dataset and conducted a Projected Gradient Descent (PGD) attack \citep{madrytowards} with varying strengths. 

Figure \ref{fig:adv} illustrates that all the SSL models trained with priors are considerably more robust against adversarial attacks of different strengths. The substantial gap in performance suggests that shape awareness assists in learning more generic features that are less vulnerable to adversarial perturbations. Hence, the proposed SSL-Prior approach can offer an effective solution for learning representations that are both accurate and robust.


\subsection{Texture Bias}

Texture bias is a well-known limitation of DNNs, where they tend to rely excessively on local texture cues in the data. We conduct a texture bias analysis by applying style-transfer \citep{huang2017arbitrary} of varying strengths on TinyImageNet images. The stylized images have different textures, and the model's ability to infer well on them indicates its ability to learn beyond local texture cues and capture the global structure of objects. We evaluate the model pre-trained on TinyImageNet on this stylized dataset that has varying textures.

Figure \ref{fig:combined} (a) shows the performances across the strength spectrum. SSL-Prior consistently performs better, thus exhibiting lower texture bias. The added shape awareness guides the model to learn global structures instead of only the local textures, thus proving robust to textural changes in challenging applications.

\subsection{Robustness against Corruptions}

DNNs have shown vulnerability to natural corruptions such as variations in input data due to changes in weather, lighting, and digital effects. Such inconsistencies in predictions can have detrimental effects, particularly in safety-critical applications. In order to evaluate the robustness of SSL-Prior models to these corruptions, we simulated a dataset by introducing fifteen different types of corruptions on the CIFAR10 dataset \citep{hendrycksbenchmarking}, grouped into four categories.

Figure \ref{fig:combined} (b) shows that SSL-Prior exhibits greater resistance to all categories of corruptions, namely blur, digital, noise, and weather, thereby demonstrating improved robustness in complex real-world situations. In challenging conditions such as noisy and changing weather, shape awareness plays a particularly significant role in enhancing the robustness of the SSL-Prior approach. This suggests that SSL with prior is better suited for real-time applications.



\subsection{Data Scarcity}

In the age when labeled data is limited and expensive, achieving comparable results with a reduced dataset is an attractive solution. To evaluate the effectiveness of SSL prior to such scenarios, we trained SSL models using varying percentages of training data, ranging from $10\%$ to $100\%$.
Figure \ref{fig:combined}(c) shows that models trained with prior knowledge consistently outperform the SSL baseline models across all percentages of training data. 
Therefore, this approach can prove particularly advantageous in scenarios where labeled data is scarce, a common predicament across numerous real-world scenarios.

\subsection{Object Detection}


\begin{table}[t]
\caption{SSL-Prior outperforms SSL baselines on the downstream detection and segmentation tasks trained with SSD detector.}
\label{tbl:det}
\begin{center}
\begin{tabular}{c|c|c}
\toprule
 Task & \multicolumn{2}{c}{Detection (mAP)} \\ \midrule
 Dataset & VOC & COCO \\ \midrule
SSL & 40.21 &15.68 \\
SSL-Prior & \textbf{56.79 } &\textbf{19.14} \\
\bottomrule
\end{tabular}
\end{center}
\end{table}

Object detection involves the classification and localization of objects in a scene and has widespread applications in various fields such as autonomous driving, surveillance, healthcare, and robotics \citep{arani2022a}. Object detection is a common downstream task, and SSL has the potential to significantly improve the performance of detection models and reduce the need for extensive labeled data. 

We evaluated the effectiveness of our proposed SSL-Prior approach on downstream tasks using pre-trained SSL models as encoders for object detection on the PASCAL VOC dataset \citep{PASCALVOC} and MS COCO datasets \citep{lin2014microsoft} with the SSD object detector \citep{SSD}. The mean average precision (mAP) is reported in Table \ref{tbl:det}. Prior-induced SSL substantially accelerates the performance of downstream tasks, as illustrated by the $ 40\%$ improvement in mAP on VOC and $ 22\%$ on COCO. Additionally, we trained detection models using encoders trained with the full set of SSL augmentations and observed a $10\%$ gain (Appendix), further highlighting the effectiveness of the SSL-Prior approach for challenging downstream tasks. This alternative training approach of SSL not only improves the performance and robustness but also demonstrates significant improvement in challenging downstream dense prediction tasks.

\section{Conclusion}
The dependencies of Self-supervised learning (SSL) training on design and considerations can have undesirable effect on applications in various domains. SSL-Prior, introduces a transformative direction by incorporating prior knowledge to emphasize global structures and foster biologically plausible representations. SSL-Prior reduces the reliance on an extensive set of augmentations, while also minimizing shortcut learning and texture bias behaviors. Additionally, the models also display strong resistance to both natural and adversarial corruptions, making it a promising technique for improving both the quality and effectiveness of learned representations. Through substantial improvements in challenging dense prediction tasks, our method exemplifies the potential of alternate approaches. This work paves the way for enhanced SSL training, offering avenue to cultivate more effective and versatile learned representations with broader applications.



\bibliography{collas2024_conference}

\begin{thebibliography}{33}
\providecommand{\natexlab}[1]{#1}
\providecommand{\url}[1]{\texttt{#1}}
\expandafter\ifx\csname urlstyle\endcsname\relax
  \providecommand{\doi}[1]{doi: #1}\else
  \providecommand{\doi}{doi: \begingroup \urlstyle{rm}\Url}\fi

\bibitem[Arani et~al.(2022)Arani, Gowda, Mukherjee, Magdy, Kathiresan, and
  Zonooz]{arani2022a}
Elahe Arani, Shruthi Gowda, Ratnajit Mukherjee, Omar Magdy,
  Senthilkumar~Sockalingam Kathiresan, and Bahram Zonooz.
\newblock A comprehensive study of real-time object detection networks across
  multiple domains: A survey.
\newblock \emph{Transactions on Machine Learning Research}, 2022.
\newblock ISSN 2835-8856.
\newblock URL \url{https://openreview.net/forum?id=ywr5sWqQt4}.
\newblock Survey Certification.

\bibitem[Bai et~al.(2022)Bai, Yang, Wang, Du, Han, Deng, Wang, and Liu]{bairsa}
Yingbin Bai, Erkun Yang, Zhaoqing Wang, Yuxuan Du, Bo~Han, Cheng Deng, Dadong
  Wang, and Tongliang Liu.
\newblock Rsa: Reducing semantic shift from aggressive augmentations for
  self-supervised learning.
\newblock In \emph{Advances in Neural Information Processing Systems}, 2022.

\bibitem[Balestriero et~al.(2022)Balestriero, Bottou, and
  LeCun]{balestriero2022effects}
Randall Balestriero, Leon Bottou, and Yann LeCun.
\newblock The effects of regularization and data augmentation are class
  dependent.
\newblock \emph{arXiv preprint arXiv:2204.03632}, 2022.

\bibitem[Bardes et~al.(2022)Bardes, Ponce, and LeCun]{bardesvicreg}
Adrien Bardes, Jean Ponce, and Yann LeCun.
\newblock Vicreg: Variance-invariance-covariance regularization for
  self-supervised learning.
\newblock In \emph{10th International Conference on Learning Representations,
  ICLR 2022}, 2022.

\bibitem[Chen et~al.(2020{\natexlab{a}})Chen, Kornblith, Norouzi, and
  Hinton]{chen2020simple}
Ting Chen, Simon Kornblith, Mohammad Norouzi, and Geoffrey Hinton.
\newblock A simple framework for contrastive learning of visual
  representations.
\newblock In \emph{International conference on machine learning}, pp.\
  1597--1607. PMLR, 2020{\natexlab{a}}.

\bibitem[Chen et~al.(2021)Chen, Luo, and Li]{chen2021intriguing}
Ting Chen, Calvin Luo, and Lala Li.
\newblock Intriguing properties of contrastive losses.
\newblock \emph{Advances in Neural Information Processing Systems}, 34, 2021.

\bibitem[Chen \& He(2021)Chen and He]{chen2021exploring}
Xinlei Chen and Kaiming He.
\newblock Exploring simple siamese representation learning.
\newblock In \emph{Proceedings of the IEEE/CVF Conference on Computer Vision
  and Pattern Recognition}, pp.\  15750--15758, 2021.

\bibitem[Chen et~al.(2020{\natexlab{b}})Chen, Fan, Girshick, and
  He]{chen2020improved}
Xinlei Chen, Haoqi Fan, Ross Girshick, and Kaiming He.
\newblock Improved baselines with momentum contrastive learning.
\newblock \emph{arXiv preprint arXiv:2003.04297}, 2020{\natexlab{b}}.

\bibitem[Everingham et~al.(2010)Everingham, Van~Gool, Williams, Winn, and
  Zisserman]{PASCALVOC}
Mark Everingham, Luc Van~Gool, Christopher K.~I. Williams, John Winn, and
  Andrew Zisserman.
\newblock The pascal visual object classes (voc) challenge.
\newblock \emph{International Journal of Computer Vision}, 88\penalty0
  (2):\penalty0 303--338, Jun 2010.
\newblock ISSN 1573-1405.
\newblock \doi{10.1007/s11263-009-0275-4}.
\newblock URL \url{https://doi.org/10.1007/s11263-009-0275-4}.

\bibitem[Geirhos et~al.(2019)Geirhos, Rubisch, Michaelis, Bethge, Wichmann, and
  Brendel]{geirhos2019imagenettrained}
Robert Geirhos, Patricia Rubisch, Claudio Michaelis, Matthias Bethge, Felix~A.
  Wichmann, and Wieland Brendel.
\newblock Imagenet-trained cnns are biased towards texture; increasing shape
  bias improves accuracy and robustness, 2019.

\bibitem[Geirhos et~al.(2020)Geirhos, Jacobsen, Michaelis, Zemel, Brendel,
  Bethge, and Wichmann]{geirhos2020shortcut}
Robert Geirhos, J{\"o}rn-Henrik Jacobsen, Claudio Michaelis, Richard Zemel,
  Wieland Brendel, Matthias Bethge, and Felix~A Wichmann.
\newblock Shortcut learning in deep neural networks.
\newblock \emph{Nature Machine Intelligence}, 2\penalty0 (11):\penalty0
  665--673, 2020.

\bibitem[Gidaris et~al.(2018)Gidaris, Singh, and
  Komodakis]{gidarisunsupervised}
Spyros Gidaris, Praveer Singh, and Nikos Komodakis.
\newblock Unsupervised representation learning by predicting image rotations.
\newblock In \emph{ICLR 2018}, 2018.

\bibitem[Gowda et~al.(2022)Gowda, Zonooz, and Arani]{gowda2022inbiased}
Shruthi Gowda, Bahram Zonooz, and Elahe Arani.
\newblock Inbiased: Inductive bias distillation to improve generalization and
  robustness through shape-awareness.
\newblock In \emph{Conference on Lifelong Learning Agents}, pp.\  1026--1042.
  PMLR, 2022.

\bibitem[Grill et~al.(2020)Grill, Strub, Altch{\'e}, Tallec, Richemond,
  Buchatskaya, Doersch, Avila~Pires, Guo, Gheshlaghi~Azar,
  et~al.]{grill2020bootstrap}
Jean-Bastien Grill, Florian Strub, Florent Altch{\'e}, Corentin Tallec, Pierre
  Richemond, Elena Buchatskaya, Carl Doersch, Bernardo Avila~Pires, Zhaohan
  Guo, Mohammad Gheshlaghi~Azar, et~al.
\newblock Bootstrap your own latent-a new approach to self-supervised learning.
\newblock \emph{Advances in Neural Information Processing Systems},
  33:\penalty0 21271--21284, 2020.

\bibitem[He et~al.(2016)He, Zhang, Ren, and Sun]{he2016deep}
Kaiming He, Xiangyu Zhang, Shaoqing Ren, and Jian Sun.
\newblock Deep residual learning for image recognition.
\newblock In \emph{Proceedings of the IEEE conference on computer vision and
  pattern recognition}, pp.\  770--778, 2016.

\bibitem[He et~al.(2020)He, Fan, Wu, Xie, and Girshick]{he2020momentum}
Kaiming He, Haoqi Fan, Yuxin Wu, Saining Xie, and Ross Girshick.
\newblock Momentum contrast for unsupervised visual representation learning.
\newblock In \emph{Proceedings of the IEEE/CVF conference on computer vision
  and pattern recognition}, pp.\  9729--9738, 2020.

\bibitem[Hendrycks \& Dietterich(2018)Hendrycks and
  Dietterich]{hendrycksbenchmarking}
Dan Hendrycks and Thomas Dietterich.
\newblock Benchmarking neural network robustness to common corruptions and
  perturbations.
\newblock In \emph{International Conference on Learning Representations}, 2018.

\bibitem[Hendrycks et~al.(2021)Hendrycks, Basart, Mu, Kadavath, Wang, Dorundo,
  Desai, Zhu, Parajuli, Guo, et~al.]{hendrycks2021many}
Dan Hendrycks, Steven Basart, Norman Mu, Saurav Kadavath, Frank Wang, Evan
  Dorundo, Rahul Desai, Tyler Zhu, Samyak Parajuli, Mike Guo, et~al.
\newblock The many faces of robustness: A critical analysis of
  out-of-distribution generalization.
\newblock In \emph{Proceedings of the IEEE/CVF International Conference on
  Computer Vision}, pp.\  8340--8349, 2021.

\bibitem[Hinton et~al.(2015)Hinton, Vinyals, and Dean]{hinton2015distilling}
Geoffrey Hinton, Oriol Vinyals, and Jeff Dean.
\newblock Distilling the knowledge in a neural network.
\newblock \emph{arXiv preprint arXiv:1503.02531}, 2015.

\bibitem[Huang \& Belongie(2017)Huang and Belongie]{huang2017arbitrary}
Xun Huang and Serge Belongie.
\newblock Arbitrary style transfer in real-time with adaptive instance
  normalization.
\newblock In \emph{Proceedings of the IEEE international conference on computer
  vision}, pp.\  1501--1510, 2017.

\bibitem[Jain et~al.(2022)Jain, Tsipras, and Madry]{jain2022combining}
Saachi Jain, Dimitris Tsipras, and Aleksander Madry.
\newblock Combining diverse feature priors.
\newblock In \emph{International Conference on Machine Learning}, pp.\
  9802--9832. PMLR, 2022.

\bibitem[Kuhl(2000)]{kuhl2000new}
Patricia~K Kuhl.
\newblock A new view of language acquisition.
\newblock \emph{Proceedings of the National Academy of Sciences}, 97\penalty0
  (22):\penalty0 11850--11857, 2000.

\bibitem[Lin et~al.(2014)Lin, Maire, Belongie, Hays, Perona, Ramanan,
  Doll{\'a}r, and Zitnick]{lin2014microsoft}
Tsung-Yi Lin, Michael Maire, Serge Belongie, James Hays, Pietro Perona, Deva
  Ramanan, Piotr Doll{\'a}r, and C~Lawrence Zitnick.
\newblock Microsoft coco: Common objects in context.
\newblock In \emph{Computer Vision--ECCV 2014: 13th European Conference,
  Zurich, Switzerland, September 6-12, 2014, Proceedings, Part V 13}, pp.\
  740--755. Springer, 2014.

\bibitem[Liu et~al.(2016)Liu, Anguelov, Erhan, Szegedy, Reed, Fu, and
  Berg]{SSD}
Wei Liu, Dragomir Anguelov, Dumitru Erhan, Christian Szegedy, Scott Reed,
  Cheng-Yang Fu, and Alexander~C Berg.
\newblock Ssd: Single shot multibox detector.
\newblock In \emph{European conference on computer vision}, pp.\  21--37.
  Springer, 2016.

\bibitem[Liu et~al.(2015)Liu, Luo, Wang, and Tang]{liu2015deep}
Ziwei Liu, Ping Luo, Xiaogang Wang, and Xiaoou Tang.
\newblock Deep learning face attributes in the wild.
\newblock In \emph{Proceedings of the IEEE international conference on computer
  vision}, pp.\  3730--3738, 2015.

\bibitem[Madry et~al.(2018)Madry, Makelov, Schmidt, Tsipras, and
  Vladu]{madrytowards}
Aleksander Madry, Aleksandar Makelov, Ludwig Schmidt, Dimitris Tsipras, and
  Adrian Vladu.
\newblock Towards deep learning models resistant to adversarial attacks.
\newblock In \emph{International Conference on Learning Representations}, 2018.

\bibitem[Minderer et~al.(2020)Minderer, Bachem, Houlsby, and
  Tschannen]{minderer2020automatic}
Matthias Minderer, Olivier Bachem, Neil Houlsby, and Michael Tschannen.
\newblock Automatic shortcut removal for self-supervised representation
  learning.
\newblock In \emph{International Conference on Machine Learning}, pp.\
  6927--6937. PMLR, 2020.

\bibitem[Mishra et~al.(2020)Mishra, Shah, Bansal, Choi, Shrivastava, Sharma,
  and Jacobs]{mishra2020learning}
Shlok Mishra, Anshul Shah, Ankan Bansal, Jonghyun Choi, Abhinav Shrivastava,
  Abhishek Sharma, and David Jacobs.
\newblock Learning visual representations for transfer learning by suppressing
  texture.
\newblock \emph{arXiv preprint arXiv:2011.01901}, 2020.

\bibitem[Noroozi \& Favaro(2016)Noroozi and Favaro]{noroozi2016unsupervised}
Mehdi Noroozi and Paolo Favaro.
\newblock Unsupervised learning of visual representations by solving jigsaw
  puzzles.
\newblock In \emph{Computer Vision--ECCV 2016: 14th European Conference,
  Amsterdam, The Netherlands, October 11-14, 2016, Proceedings, Part VI}, pp.\
  69--84. Springer, 2016.

\bibitem[Pearl \& Mackenzie(2018)Pearl and Mackenzie]{pearl2018book}
Judea Pearl and Dana Mackenzie.
\newblock \emph{The book of why: the new science of cause and effect}.
\newblock Basic books, 2018.

\bibitem[Robinson et~al.(2021)Robinson, Sun, Yu, Batmanghelich, Jegelka, and
  Sra]{robinson2021can}
Joshua Robinson, Li~Sun, Ke~Yu, Kayhan Batmanghelich, Stefanie Jegelka, and
  Suvrit Sra.
\newblock Can contrastive learning avoid shortcut solutions?
\newblock \emph{Advances in Neural Information Processing Systems}, 34, 2021.

\bibitem[Xiao et~al.(2021)Xiao, Engstrom, Ilyas, and Madry]{xiao2021noise}
Kai Xiao, Logan Engstrom, Andrew Ilyas, and Aleksander Madry.
\newblock Noise or signal: The role of image backgrounds in object recognition.
\newblock In \emph{International Conference on Learning Representations}, 2021.

\bibitem[Zbontar et~al.(2021)Zbontar, Jing, Misra, LeCun, and
  Deny]{zbontar2021barlow}
Jure Zbontar, Li~Jing, Ishan Misra, Yann LeCun, and St{\'e}phane Deny.
\newblock Barlow twins: Self-supervised learning via redundancy reduction.
\newblock In \emph{International Conference on Machine Learning}, pp.\
  12310--12320. PMLR, 2021.

\end{thebibliography}
\bibliographystyle{collas2024_conference}

\newpage
\appendix
\section{Appendix}


\subsection{Shortcut Learning}
\label{sec:short_app}
We conducted two experiments to examine shortcut learning behavior, using Tinted-STL10 and Skewed-CelebA datasets. We used SSL pre-trained models from the TinyImageNet dataset, and fine-tuned them for evaluation on the specific datasets. Figure \ref{fig:shortcut} showed the comparison with the SSL baselines that were trained with basic augmentations. Here, we report the results of the SSL baseline (VICReg) with a full set of augmentations (Strong Augmentations in Table \ref{tbl:aug}). 

\begin{figure}[h] 
    \centering
    \includegraphics[width=0.55\textwidth]{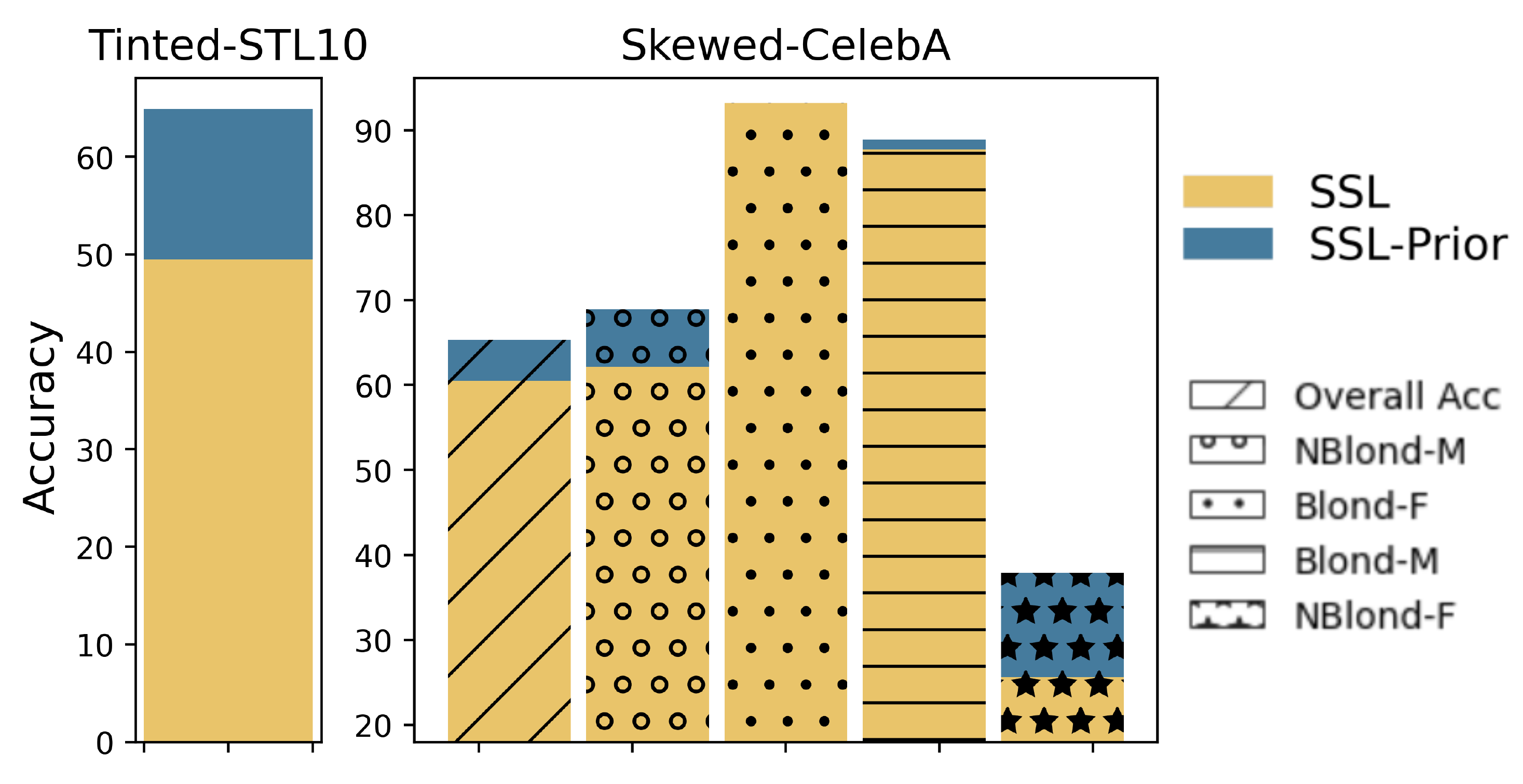} 
    \caption{Shortcut learning: Evaluation using Tinted-STL10 and Skewed-CelebA datasets on all VICReg SSL method, with \textit{strong augmentations}.}
    \label{fig:shortut2}%
\end{figure}

In the Tinted-STL10 experiment, we introduced a class-specific tint to each training sample in the STL10 dataset and fine-tuned the classifier on this modified dataset. SSL training, which learned the "tint" shortcut in the sample, did not perform well on the test set since it did not contain this irrelevant correlation. It can be seen in Figure \ref{fig:shortut2} that the prior induced SSL is less vulnerable to learning shortcuts, even compared to the SSL baseline that utilized all augmentations.

In Skewed-CelebA, the training data consisted solely of images of blonde females and non-blonde males. However, the test set also includes images of the other two categories: non-blonde females and blonde males. Figure \ref{fig:shortut2} shows the accuracy scores for each category, as well as the overall accuracy. The SSL baseline exhibited lower accuracy in the non-blonde female category, as the training had associated blonde hair with the female gender. However, the use of prior knowledge eliminated this bias and resulted in a more generalizable approach.

\subsection{Ablation Study}
\label{sec:abl_app}

Highlighting the significance of specific inductive bias, particularly its introduction into the training paradigm, remains a pivotal aspect of our study. To underscore this importance, we conduct additional ablation experiments. Training exclusively on shape images or simply incorporating shape as an augmentation strategy yields suboptimal features, mainly due to the inherent distribution discrepancies. Knowledge distillation \citep{hinton2015distilling} distills the knowledge from one network to another. We leverage the concept of knowledge distillation but in the form of self-teaching \citep{gowda2022inbiased} between different networks. This approach serves as a key mechanism to incorporate shape-inductive bias into the SSL training framework.

To further elucidate this principle, we present additional experimental results. We train a baseline SimSiam model while incorporating the shape as an augmentation, as detailed in Table \ref{tbl:base}. In particular, introducing the Shape filter as an augmentation into the SimSiam baseline results in a notable performance decrease (from 36.78 to $29.65$). This illustrates that just appending to an augmentation list is suboptimal. The guidance imparted by another network concurrently learning on shape features serves as a catalyst, steering the primary network towards learning more generalizable features. Second, we also analyze the impact of inducing any other common augmentation (or filter) as the prior instead of shape. Hence, we employ a common augmentation technique, specifically color distortion, as a form of prior knowledge to train the secondary network. Subsequent to this training, we observe an improvement in the performance of the network, with a score of $52.44$. However, it is noteworthy that this enhancement remains notably inferior to the results obtained when employing shape as the primary prior, thereby accentuating the pivotal role of shape features in our framework.

These two experiments collectively establish that shape as an inductive bias improves the generalization capability. Moreover, they underscore the critical significance of the manner in which this knowledge is induced, demonstrating its pivotal role in improving the overall training process.

\begin{figure}[h] 
    \centering
    \includegraphics[width=0.6\textwidth]{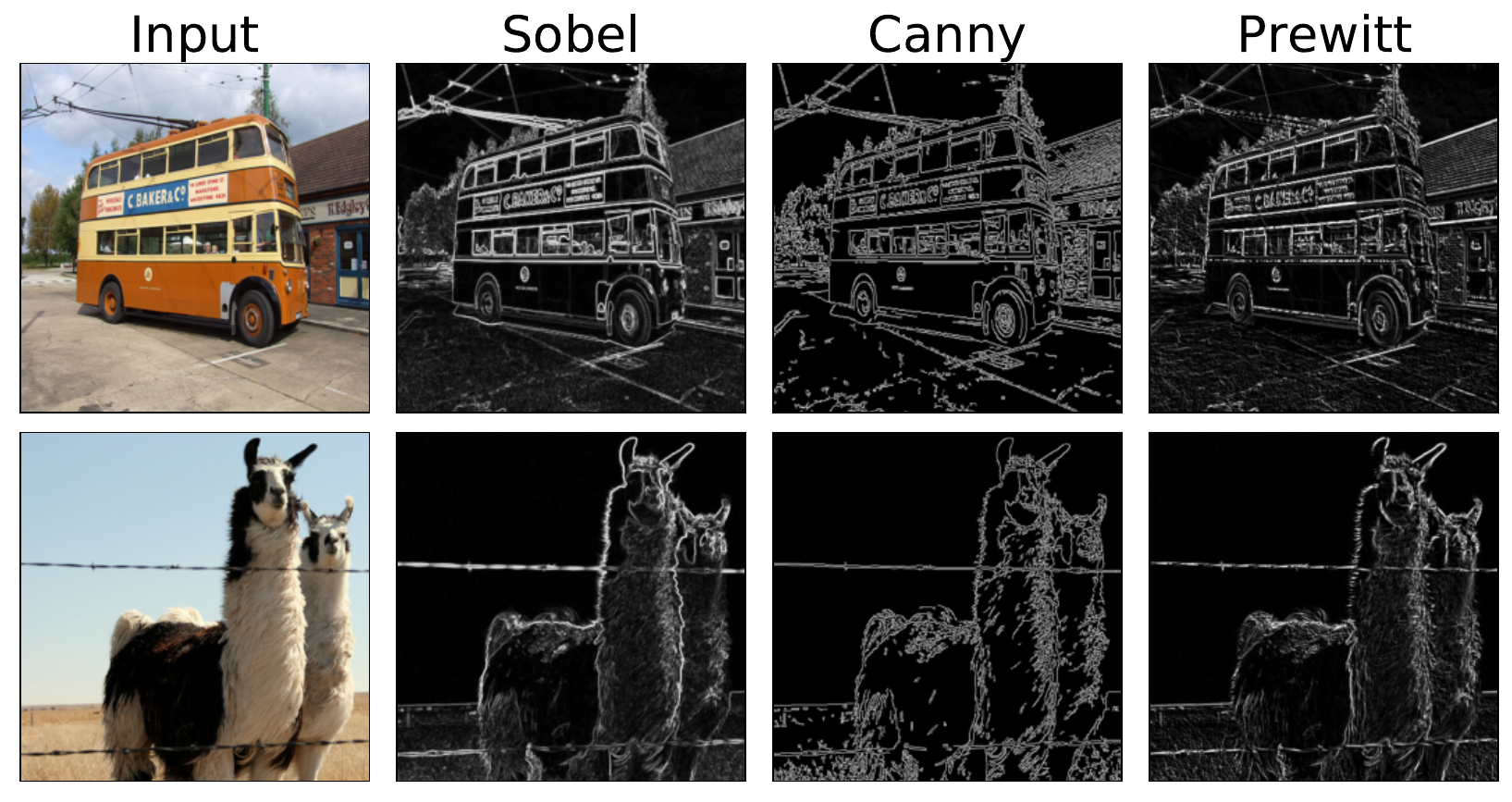} 
    \caption{Sample images of different shape filters}
    \label{fig:shape}%
\end{figure}

Furthermore, we also show the effect of using different shape filters to extract shape information in Table \ref{tbl:base}. The Canny filter produces more noisy edges compared to the others, and the Sobel filter performs the best. Sample images are shown in Figure \ref{fig:shape}
.
\begin{table}[h]
\caption{Additional results and ablations.}
\label{tbl:base}
\begin{center}
\begin{tabular}{l|c}
\hline
Method & CIFAR100 \\ \hline
SimSiam & 36.78 \\
SimSiam - w/ shape as augm. & 29.65 \\
SimSiam - w/ network(color)  & 52.44   \\ \hline
SimSiam-Prior (Sobel)   & \textbf{66.06}  \\
SimSiam-Prior (Canny)   & 44.03 \\
SimSiam-Prior (Prewitt) & 55.62  \\
\hline
\end{tabular}
\end{center}
\end{table}

\subsection{Object Detection}

Object detection is a dense prediction task that is used in a multitude of applications. Table \ref{tbl:det} showed the results of the SSL baselines and the SSL-Prior framework when using basic augmentations. To assess the effectiveness of the downstream performance of SSL models trained with the extensive set of augmentations, we train the SSL model (VICReg) with strong augmentations and use this model for training the object detection task. For object detection, we employed an encoder pre-trained on the TinyImageNet dataset, in addition to the SSD detection head.

Table \ref{tbl:det2} shows that the SSL model trained with basic augmentations and prior knowledge induction outperforms both SSL baselines. SSL training with prior knowledge aids in identifying salient object features, which, in turn, improves the classification and localization of different objects in the image.


\begin{table}[h]
\caption{SSL-Prior outperforms SSL baselines on the downstream detection task trained with SSD detector on VOC dataset.}
\label{tbl:det2}
\begin{center}
\begin{tabular}{l|ccc}
\toprule
 & \begin{tabular}[c]{@{}c@{}}SSL\\ (Basic Aug)\end{tabular} & \begin{tabular}[c]{@{}c@{}}SSL\\ (Strong Aug)\end{tabular} & SSL-Prior \\ \midrule
Object Detection & 40.21 & 51.89 & \textbf{56.79} \\ \bottomrule
\end{tabular}
\end{center}
\end{table}

\subsection{Strong Augmentations}

We also report SSL baselines with strong augmentations (the original set of augmentations for each method reported in the literature) and tabulate the results in Table \ref{tbl:acc2}. SSL with prior knowledge is not only robust to shortcut learning, texture bias, and corruptions, but also achieves comparable IID classification accuracies.

\begin{table}[h]
\caption{The SSL baselines trained with \textit{Strong augmentations}.}
\label{tbl:acc2}
\begin{center}
\begin{tabular}{l|cccc}
\toprule
& CIFAR10 & CIFAR100 & STL10  & TinyImageNet  \\ 
\midrule
SimCLR  & 89.75 & 61.30 & 85.10 & 39.40 \\
SimSiam & 89.90 & 59.28 & 81.29 & 31.54  \\
VICReg & 90.05 & 66.30 & 82.16 & 42.72  \\
\bottomrule
\end{tabular}
\end{center}
\end{table}

\begin{table}[t]
\caption{Hyperparameters for SSL baselines and also the SSL-Prior counterparts.}
\label{tbl:hyper}
\begin{center}
\begin{small}
\begin{tabular}{l|l|l} 
\toprule
Dataset & SimCLR & SimCLR-Prior \\ \midrule
\begin{tabular}[c]{@{}l@{}}CIFAR10\\ CIFAR100\\ STL10\end{tabular} & 
\begin{tabular}[c]{@{}l@{}}lr: 0.0003\\ weight decay: 1e-6 \\ epochs: 1000\\ batch size: 512\\ proj layer: 2048\end{tabular} & 
\begin{tabular}[c]{@{}l@{}}lr: 0.0003\\ weight decay: 1e-6 \\ epochs: 1000\\ batch size: 512\\ proj layer: 2048 \\ $\lambda$: 0.5\end{tabular} \\ 
\midrule
TinyImageNet & 
\begin{tabular}[c]{@{}l@{}}lr: 0.0002\\ weight decay: 1e-6 \\ epochs: 1000\\ batch size: 512\\ proj layer: 2048\end{tabular}  & 
\begin{tabular}[c]{@{}l@{}}lr: 0.0002\\ weight decay: 1e-6 \\ epochs: 1000\\ batch size: 512\\ proj layer: 2048 \\ $\lambda$: 1.0\end{tabular} \\ 
\midrule

Dataset & SimSiam & SimSiam-Prior \\ \midrule
\begin{tabular}[c]{@{}l@{}}CIFAR10\\ CIFAR100\\ STL10\end{tabular} & 
\begin{tabular}[c]{@{}l@{}}lr: 0.0003\\ weight decay: 1e-6 \\ epochs: 1000\\ batch size: 256\\ proj layer: 2048 \\ pred\_layer: 512\end{tabular} & 
\begin{tabular}[c]{@{}l@{}}lr: 0.0003\\ weight decay: 1e-6 \\ epochs: 1000\\ batch size: 256\\ proj layer: 2048 \\ pred\_layer: 512\\ $\lambda$: 1.0\end{tabular} \\ 
\midrule
TinyImageNet & 
\begin{tabular}[c]{@{}l@{}}lr: 0.0002\\ weight decay: 1e-6 \\ epochs: 1000\\ batch size: 256\\ proj layer: 2048 \\ pred\_layer: 512\end{tabular} & 
\begin{tabular}[c]{@{}l@{}}lr: 0.0002\\ weight decay: 1e-6 \\ epochs: 1000\\ batch size: 256\\ proj layer: 2048 \\ pred\_layer: 512\\ $\lambda$: 1.0\end{tabular} \\ 
\midrule

Dataset & VICReg & VicReg-Prior \\ \midrule
\begin{tabular}[c]{@{}l@{}}CIFAR10\\ CIFAR100\\ STL10\end{tabular} & 
\begin{tabular}[c]{@{}l@{}}lr: 0.0003\\ weight\_decay: 1e-6 \\ epochs: 1000\\ batch size: 256 \\proj layer: 2048 \end{tabular} & 
\begin{tabular}[c]{@{}l@{}}lr: 0.0003\\ weight decay: 1e-6 \\ epochs: 1000\\ batch size: 256\\ proj layer: 2048 \\ $\lambda$: 2.0\end{tabular} \\ 
\midrule
TinyImageNet & 
\begin{tabular}[c]{@{}l@{}}lr: 0.0002\\ weight decay: 1e-6 \\ epochs: 200\\ batch size: 256 \\proj layer: 2048\end{tabular} & 
\begin{tabular}[c]{@{}l@{}}lr: 0.0002\\ weight decay: 1e-6 \\ epochs: 200\\ batch size: 256\\ proj layer: 2048 \\ $\lambda$: 2.0\end{tabular} \\ 
\bottomrule
\end{tabular}
\end{small}
\end{center}
\end{table}

\subsection{Implementation Details}

One SSL method is chosen from the three broad categories: SimCLR from contrastive-based approaches, SimSiam from asymmetric-based approaches, and VICReg from feature decorrelation-based approaches for all experiments.

ResNet-18 architecture \citep{he2016deep} is used in all experiments. For small datasets, the modified version of ResNet-18 is used, with the $7x7$ kernel replaced with $3x3$ and without a maxpool layer. A two-layer projection MLP layer is appended to the encoder. For asymmetric methods that require a predictor, there is an additional two-layer MLP predictor. Adam Optimizer and Cosine Annealing scheduler are used for all the training. A detailed set of hyperparameters for each method is provided in Table \ref{tbl:hyper}.

\textbf{Evaluation}
All the models are evaluated using the standard linear evaluation protocol used in the SSL algorithm. For evaluation, a fine-tuning step is performed in which a linear classifier head is trained on top of the frozen encoder. The linear evaluation is performed for 100 epochs, and the weight decay is set to zero.

\subsection{Inductive Bias}
We have included the details of the prior filter in Algorithm \ref{algo:sobel}.

\begin{algorithm}[h]
   \caption{Sobel Algorithm}
   \label{algo:sobel}
   \begin{small}
\begin{algorithmic}[1]
    \Statex Input image $x_{in}$ to Prior network
    \Statex $x_{pr} = Prior(x_{in})$
    \State Up-sample the images: $x_{up}$ = UpSamp$(x_{in})$
    \State Gaussian smoothing to reduce noisy edges: \newline
    $x_{s}$ = Gaussian$(x_{up})$
    \State Sobel kernels: \newline
        $k_x = \begin{bmatrix}
                -1 & 0 & +1 \\
                -2 & 0 & +2 \\
                -1 & 0 & +1
                \end{bmatrix}$ and 
        $k_y = \begin{bmatrix}
                -1 & -2 & -1 \\
                 0 &  0 &  0 \\
                +1 & +2 & +1
                \end{bmatrix}$
    \State $x_{sx} = x_{s} \ast k_x$ and $x_{sy} = x_{s} \ast k_y$; \newline
    \Comment{$\ast$ is 2-dimensional convolution operation}
    \State Magnitude of edges: $x_{pr} = \sqrt{x_{sx}^2 + x_{sy}^2}$
    \State Down-sample to original image size: \newline
    $x_{pr}$ = DownSamp$(x_{pr})$
\end{algorithmic}
\end{small}
\end{algorithm}

\end{document}